\documentclass[10pt,twocolumn,letterpaper]{article}
\pdfoutput=1
\usepackage[pagenumbers]{cvpr}

\usepackage[dvipsnames]{xcolor}

\definecolor{cvprblue}{rgb}{0.21,0.49,0.74}
\usepackage[pagebackref,breaklinks,colorlinks,citecolor=cvprblue]{hyperref}

\usepackage{algorithm}
\usepackage{algorithmic}
\usepackage{amsmath}
\usepackage{amssymb}
\usepackage{booktabs}
\usepackage{bm}

\usepackage[whole]{bxcjkjatype}
\usepackage{adjustbox}
\usepackage{multirow}
\usepackage{mathtools}
\usepackage{capt-of}
\usepackage{cuted}
\usepackage{xfrac}
\usepackage{microtype}
\usepackage{colortbl}
\usepackage{xcolor}
\usepackage{makecell}
\usepackage{pifont}
\usepackage{transparent}
\usepackage{float}
\usepackage{caption}
\usepackage{subcaption}
\usepackage{graphicx}

\makeatletter
\DeclareRobustCommand\onedot{\futurelet\@let@token\@onedot}
\def\@onedot{\ifx\@let@token.\else.\null\fi\xspace}

\makeatother

\newcommand{\ours}{ResEnsemble-DDPM }
\renewcommand{\paragraph}{\vspace{1mm}\noindent\textbf}

\title{ResEnsemble-DDPM: Residual Denoising Diffusion Probabilistic Models for Ensemble Learning}

\author{
    Zhenning Shi\textsuperscript{1}\thanks{Equal contribution}\qquad
    Changsheng Dong\textsuperscript{1}\footnotemark[1]\qquad
    Xueshuo Xie\textsuperscript{2}\qquad 
    Bin Pan\textsuperscript{3}\qquad \\
    Along He\textsuperscript{1}\qquad
    Tao Li\textsuperscript{1,2}\thanks{Corresponding author: litao@nankai.edu.cn}\qquad
    \\
    \textsuperscript{1}College of Computer, Nankai University \\
    \textsuperscript{2}Haihe Lab of ITAI\\
    \textsuperscript{3}School of Statistics and Data Science, Nankai University \\
}

\begin{document}
\maketitle

\begin{abstract}
Nowadays, denoising diffusion probabilistic models have been adapted for many image segmentation tasks. However, existing end-to-end models have already demonstrated remarkable capabilities. Rather than using denoising diffusion probabilistic models alone, integrating the abilities of both denoising diffusion probabilistic models and existing end-to-end models can better improve the performance of image segmentation.
Based on this, we implicitly introduce residual term into the diffusion process and propose \ours, which seamlessly integrates the diffusion model and the end-to-end model through ensemble learning.
The output distributions of these two models are strictly symmetric with respect to the ground truth distribution, allowing us to integrate the two models by reducing the residual term. Experimental results demonstrate that our \ours can further improve the capabilities of existing models. Furthermore, its ensemble learning strategy can be generalized to other downstream tasks in image generation and get strong competitiveness.
\end{abstract}

\section{Introduction}
\label{sec:intro}

\begin{figure*}[t]
		\centering
		\includegraphics[width=0.9\linewidth]{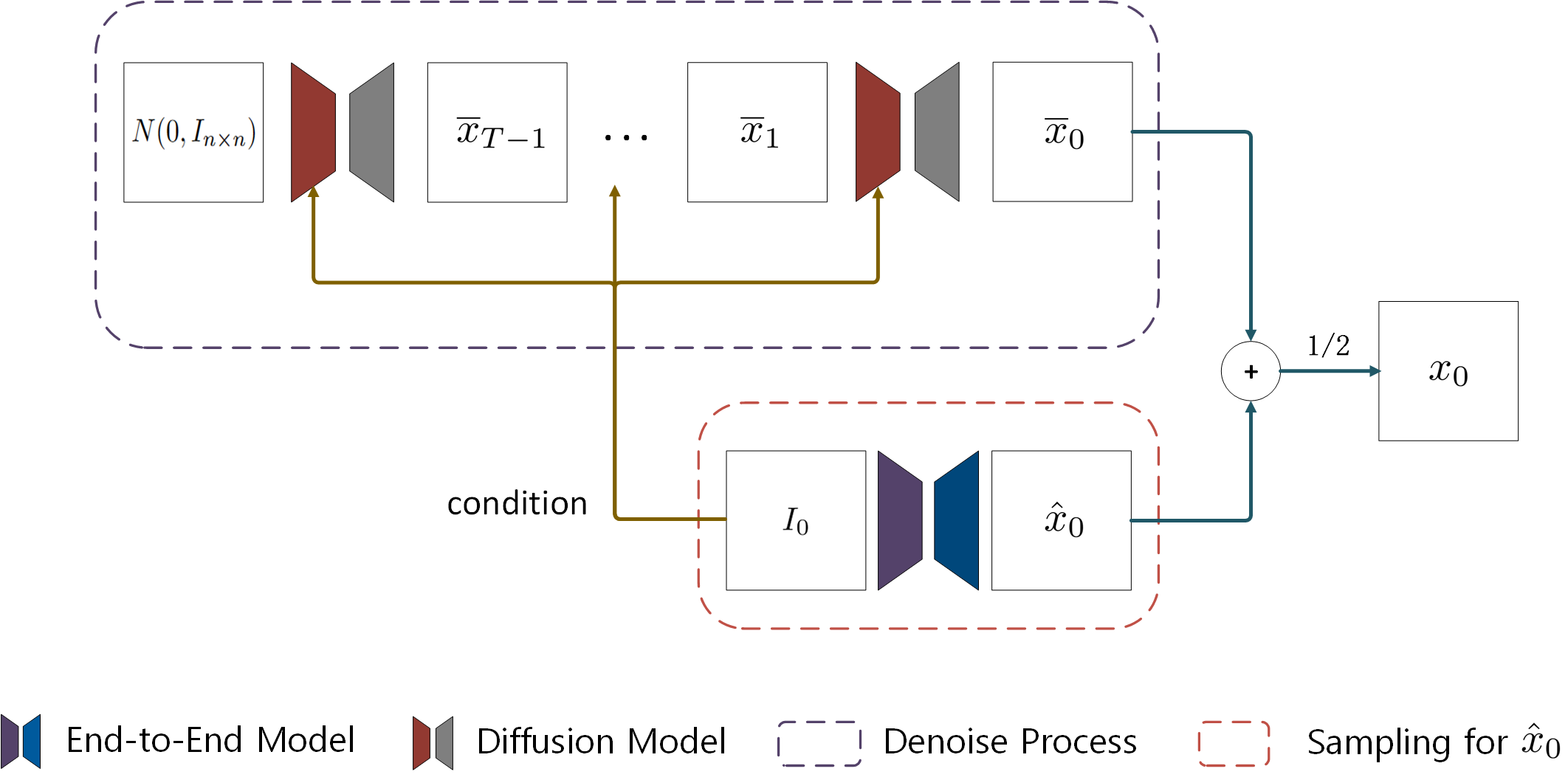}
		\caption{Inference pipeline of \ours. $\hat{x}_{0}$ represents the linkelihood output of the $E2EModel$. $\overline{x}_{0}$ is the output of the DDPM, representing the prediction of the $x_0-R$. By adding $\hat{x}_{0}$ and $\overline{x}_{0}$ together, according to Eq \ref{eq: x_0}, we can obtain the final output result of \ours.}
		\label{fig: pipeline_inference}
\end{figure*}

The mathematical modeling of the denoising diffusion models was initially proposed by \citet{sohl2015deep}. \citet{ho2020denoising} formally defined the prototype of the generalized Denoising Diffusion Probabilistic Models(DDPM). 
In contrast to the generative adversarial network\cite{goodfellow2020generative}, The denoising diffusion models generate target images in an autoregressive manner and have demonstrated remarkable competitiveness in image generation tasks, like DALL·E 2\cite{ramesh2022hierarchical}, Imagen\cite{saharia2022photorealistic} and stable diffusion\cite{rombach2022high}.\par

Due to its powerful semantic extraction capabilities, DDPM have been introduced for image segmentation. \citet{amit2021segdiff} first introduced DDPM into image segmentation tasks, following work has made various improvements base on it\cite{tian2023diffuse, wu2023diffumask, wang2023diffusion, ni2023ref, baranchuk2021label}.

However, these methods have certain limitations. First, most methods merely incorporate the diffusion model as a independent segmentation network without considering the capability of existing models. Although researchers\cite{lai2023denoising, guo2023accelerating} combine them by pre-segmentation, this method does not strictly satisfy the Markov chain and lacks theoretical proof. Secondly, these methods fail to take the residual term into account, thus preventing effective integration with existing end-to-end models. Considering these limitations, we attempt to view this problem from a higher perspective, using ensemble learning to combine the DDPM and existing end-to-end models.

In this paper, we propose \ours, a target image generation framework integrating denoising diffusion probabilistic models and existing end-to-end models by ensemble learning. Our contributions can be summarized as follows.
\begin{itemize}
    \item 
    First, we propose a residual denoising diffusion probabilistic model by adding a new distribution learned via a Markov process. This new distribution is the addition of the ground truth and the residual term, and strictly symmetric to the likelihood output distribution with respect to the ground truth.
    \item 
    Second, we seamlessly integrates denoising diffusion probabilistic models and end-to-end models by ensemble learning. By actually employ two learners with negative and positive residuals, we can effectively reduce the residual, resulting in superior segmentation capabilities compared to a single learner.
    \item Third, \ours is not restricted to the domain of image segmentation. In fact, it can be applied to various image generation tasks. It is highly adaptable as it combines both end-to-end models and denoising diffusion models. Subsequent experiments have further showed the generalization and competitive performance of \ours.
	 
\end{itemize}

Code will be released on: https://github.com/nkicsl /ResEnsemble-DDPM.

\section{Related Work}
\label{sec:related}

\textbf{Image segmentation} aims to classify pixels of an image into different categories. Various criteria can be used to describe the image segmentation tasks, mainly include semantic segmentation\cite{long2015fully}, instance segmentation\cite{he2017mask} and panoptic segmentation\cite{kirillov2019panoptic}. Semantic segmentation is to label each pixel in the image with a category label. Instance segmentation is to firstly detect the target in the image, and then label each instance of the target. Panoptic segmentation is the combination of semantic segmentation and instance segmentation, aiming to detect all targets while distinguishing different instances.

\textbf{Denoising diffusion model} is a generative model that generates target images by iterative reverse denoising process. It follows the Markov process in which the original data is progressively corrupted by noise and eventually transformed into pure Gaussian noise. The prototype of the denoising diffusion model can be traced back to the work of \citet{sohl2015deep}, where the target data was first mapped to the noise distribution. \citet{ho2020denoising}  demonstrated that this modeling approach belongs to the score-matching models, which estimate the gradient of the log-density. The score-matching models were mainly proposed and advanced by \citet{hyvarinen2005estimation}, \citet{vincent2011connection}, \citet{song2019generative, song2020improved}. This development simplifies the variational lower bound of training objective and introduces the generalized Denoising Diffusion Probabilistic Models (DDPM).

\textbf{Conditional denoising diffusion model} introduces conditions to guide the randomness of the denoising diffusion process. The unconditional denoising diffusion model learns the overall distribution, resulting in high randomness. In contrast, the conditional introduction scheme named Classifier-Guidance was initially proposed by \citet{dhariwal2021diffusion} and later extended by \citet{liu2023more}. This scheme does not require retraining the model but lacks fine-grained control. The Classifier-Free scheme was first proposed by \citet{ho2022classifier}, which has a higher implementation cost but demonstrates excellent generation control. Outstanding conditional diffusion models such as DALL·E 2\cite{ramesh2022hierarchical}, Imagen\cite{saharia2022photorealistic} and Stable Diffusion\cite{rombach2022high} are implemented based on this scheme.\par

\citet{amit2021segdiff} first introduced conditional denoise diffusion probabilistic model into image segmentation tasks. It generates segmentation masks by using the original image as a condition input to guide the denoising direction. Based on it, A series of studies have made improvements or reconstructions from different perspectives, including the architecture adjustments\cite{rahman2023ambiguous, amit2023annotator, wolleb2022diffusion, zbinden2023stochastic}, the modification of training procedure\cite{fu2023recycling, chen2022analog, fu2023importance}, the enhancements to the denoising modules\cite{wang2023dformer, xing2023diff, wu2022medsegdiff, bozorgpour2023dermosegdiff, wu2023medsegdiff, chowdary2023diffusion} and application to other downstream tasks\cite{mao2023contrastive, alimanov2023denoising, jiang2023diffused, ayala2023diffusion, ivanovska2023tomatodiff}. 

\textbf{Ensemble learning} is a machine learning approach that improves overall performance by combining predictions from multiple learners. The general structure of ensemble learning involves two or more individual learners,  combining them together through specific strategies\cite{yang2023survey}. Some methods have achieved excellent results by constructing segmentation models using ensemble learning compared to a single learner\cite{wang2015hierarchical, zheng2019new, hu2022mutual}. To fully utilize the capabilities of denoising diffusion probabilistic models and existing end-to-end models, we hope to integrate the two through ensemble learning and further improve performance.

\section{Background}
\label{sec:background}

\begin{figure*}[tp]
	\centering
	\includegraphics[width=0.9\linewidth]{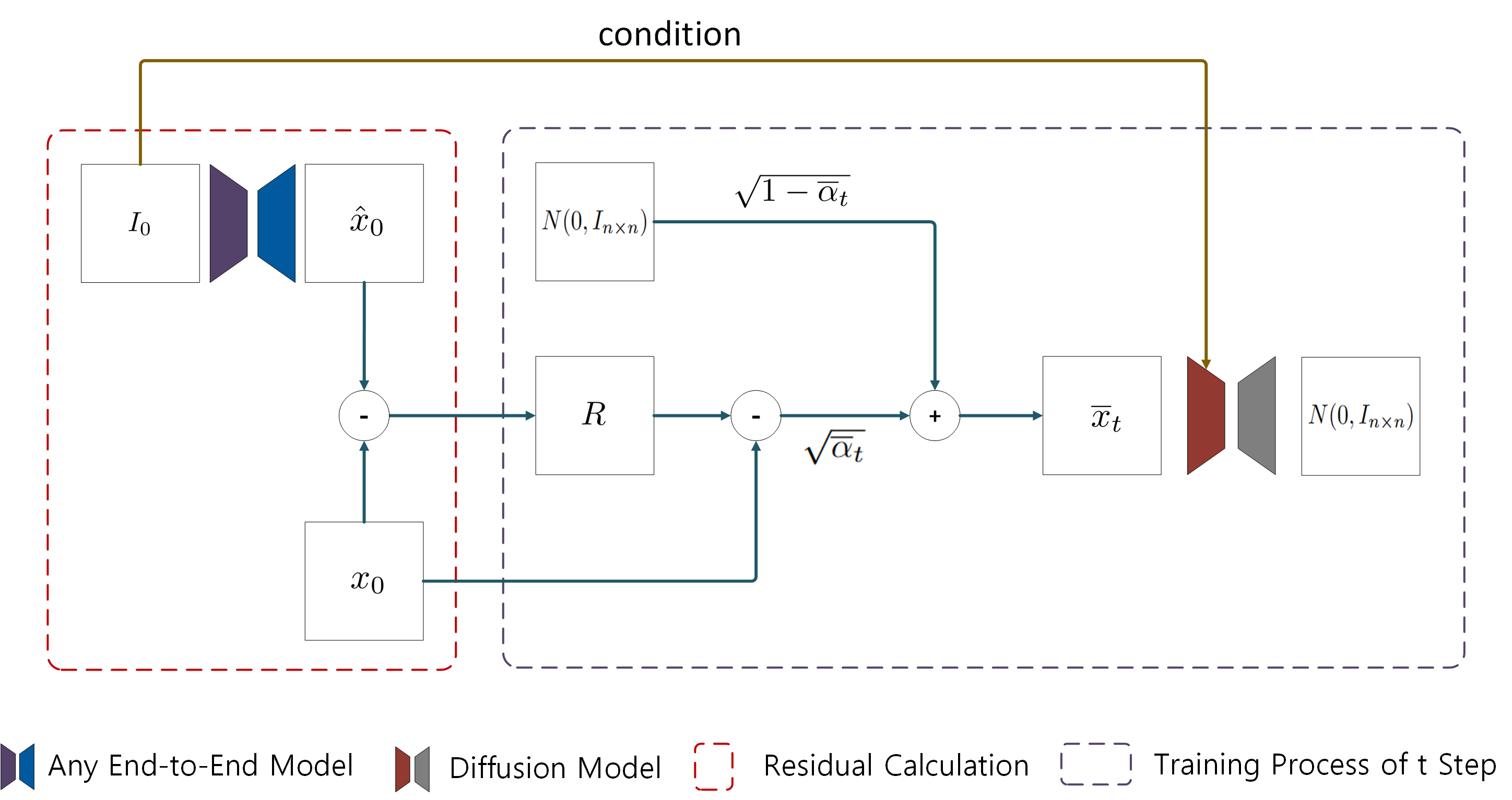}
	\caption{Training pipeline of \ours. $\hat{x}_{0}$ represents the likelihood output of the $E2EModel$. $R=\hat{x}_{0}-x_0$ represents the residual term between the likelihood output and the ground truth. We replace the original training input $x_0$ in $DDPM$ with $\overline{x}_{0}$, allowing the model to implicitly learn the distribution of $R$. We introduce $I_0$ as the conditional input.}
	\label{fig: pipeline_train}
\end{figure*}

The denoising diffusion model \cite{ho2020denoising,sohl2015deep} aims to learn a distribution :
\begin{equation}
	\label{eq:diffusion}
	P_{\theta}(x_{0}) = \int_{{x}_{1}:{x}_{T}} P_{data}({x}_{T})
	\prod_{t=1}^{T} P_{\theta}({x}_{t-1}|{x}_{t})d{x}_{1}:{x}_{T}
\end{equation}

 to approximate $P_{\theta}(x_{0})$ to the target data distribution $P_{data}(x_{0})$, where $x_{0}$ is the target image and $x_{1}, ...,x_{T}$ are latent variables with the same dimensions as $x_{0}$. In the forward process, $P_{data}(x_{0})$ is diffused into a Gaussian noise distribution through a fixed Markov chain.

\begin{equation}
	\label{eq: q_sample_Markov_chain}
	q(x_{1:T} | x_{0}) = \prod_{s=0}^{T}q(x_{t}|x_{t-1})
\end{equation}

\begin{equation}
	\label{eq: q_sample}
	q(x_{t} | x_{t-1}) = N(x_{t} ; \sqrt{1-\beta_{t}}x_{t-1} , \beta_{t}I_{n \times n})
\end{equation}

where $\beta_{t}$ is a constant that defines the level of noise, and $I_{n \times n}$ is the identity matrix of size $n$. 

\begin{equation}
	\label{eq: alpha}
	\alpha_{t} = 1 - \beta_{t},\overline{\alpha}_{t}=\prod_{s=1}^{t}\alpha_{s}
\end{equation}

As described in \cite{ho2020denoising}, ${x}_{t}$ can be reparameterized as Eq \ref{eq: reparameterized}.

\begin{equation}
	\label{eq: reparameterized}
	q(x_{t} | x_{0}) = N(x_{t} ; \sqrt{\overline\alpha_{t}}x_{0} , (1-\overline\alpha_{t})I_{n \times n})
\end{equation}

And the term to be minimized can be modified as Eq \ref{eq: DDPM_simple_minimize_term}.

\begin{equation}
	\label{eq: DDPM_simple_minimize_term}
	\mathbb{E}_{x_{0},\epsilon,t}[||\epsilon - \epsilon_{\theta}(x_{t},t)||^{2}]
\end{equation}

Then the reverse process is parameterized by $\theta$ and defined by Eq \ref{eq: DDPM_reverse}.

\begin{equation}
	\label{eq: DDPM_reverse}
	P_{\theta }(x_{0}:x_{T-1}|x_{T})=\prod_{t=1}^{T} P_{\theta}(x_{t-1}|x_{t})
\end{equation}

The reverse process converts the distribution of latent variables into the distribution of data, denoted as $P_{\theta}(x_{0})$. By taking small Gaussian steps, the reverse process is performed as Eq \ref{eq: DDPM_reverse process} and \ref{eq: DDPM_1 to 0}.

\begin{equation}
	\label{eq: DDPM_reverse process}
	P_{\theta }(x_{t-1} | x_{t}) = N(x_{t-1} ; \mu_{\theta}(x_{t},t) , \Sigma_{\theta}(x_{t},t))
\end{equation}

\begin{equation}
	\label{eq: DDPM_1 to 0}
	P_{\theta }(x_{0} | x_{1}) = N(x_{0} ; \mu_{\theta}(x_{1},1))
\end{equation}

We can obtain Eq \ref{eq: DDPM_mu_theta} according to \cite{ho2020denoising}.

\begin{equation}
	\label{eq: DDPM_mu_theta}
	\mu_{\theta}(x_t, t)=\frac{1}{\sqrt{\alpha_{t}}}(x_{t}-\frac{1-\alpha_{t}}{\sqrt{1-\overline\alpha_{t}}}\epsilon_{\theta }) 
\end{equation}

\section{Method}
\label{sec:method}


Image segmentation is a task that divides the pixels of an image into different categories. In the context of end-to-end models for image segmentation, the goal is to learn a predicted segmentation mask $\hat{x}_{0}$ for a given image $I_{0}$. By training the model on the training dataset, we can obtain the likelihood output $\hat{x}_{0}$, as shown in Eq \ref{eq:Alignment}. Here $E2EModel$ represents a pretrained end-to-end model, and $I_{0}$ represents the origin input image.

\begin{equation}
	\label{eq:Alignment}
	\hat{x}_{0} = E2EModel(I_{0})
\end{equation}

However, there is a gap between ground truth $x_{0}$ and likelihood output $\hat{x}_{0}$, since the end-to-end model cannot completely fit the ground truth. We hope to use the diffusion process to close this gap. We define the residual term as Eq \ref{eq: residual term}. 

\begin{equation}
	\label{eq: residual term}
	R = \hat{x}_{0}-x_{0}
\end{equation}

Our method is simple but solid, instead of learning the corresponding target image $x_{0}$, our learning objective becomes learning $x_{0}-R$, defined as $\overline{x}_{0}$ in Eq \ref{eq: overline_x_0}. After completing the reverse denoising process, we can obtain $\overline{x}_{0}$. According to Eq\ref{eq: hat_x_0}, we only need to add $\overline{x}_{0}$ and $\hat{x}_{0}$ together to obtain $x_{0}$ as Eq \ref{eq: x_0}, and this process is consistent with the idea of Ensemble Learning. We called these Residual Denoising Diffusion Probabilistic Models for Ensemble Learning as \textbf{\ours}.

\begin{equation}
	\label{eq: overline_x_0}
	\overline{x}_{0} = x_{0} - R
\end{equation}

\begin{equation}
	\label{eq: hat_x_0}
	\hat{x}_{0} = x_{0} + R
\end{equation}

\begin{equation}
	\label{eq: x_0}
	{x}_{0} = \frac{1}{2}(\overline{x}_{0}+\hat{x}_{0})
\end{equation}

Then we can drive the noising process similar to \cite{ho2020denoising}, as Eq \ref{eq: q_sample_one_step}.

\begin{equation}
	\label{eq: q_sample_one_step}
	q(\overline{x}_{t} | \overline{x}_{t-1}) = N(\overline{x}_{t} ; \sqrt{1-\beta_{t}}\overline{x}_{t-1} , \beta_{t}I_{n \times n})
\end{equation}

and the distribution of $\overline{x}_{t}$ can be reparameterized as Eq \ref{eq: overline_x_t_reparameterized}

\begin{equation}
	\label{eq: overline_x_t_reparameterized}
	q(\overline{x}_{t} | \overline{x}_{0}) = N(\overline{x}_{t} ; \sqrt{\overline\alpha_{t}}\overline{x}_{0} , (1-\overline\alpha_{t})I_{n \times n})
\end{equation}

Therefore, our optimization objective is the same as \cite{ho2020denoising}, which is Eq \ref{eq: Re(n)semble-DDPM_simple_minimize_term}.

\begin{equation}
	\label{eq: Re(n)semble-DDPM_simple_minimize_term}
	\mathbb{E}_{x_{0},\epsilon,t}[||\epsilon - \epsilon_{\theta}(\overline{x}_{t},t)||^{2}]
\end{equation}

Similar to the approach proposed by \citet{amit2021segdiff} in their work on image segmentation with denoising diffusion probabilistic models, our method enhances the diffusion process by incorporating a conditional function. This function integrates latent representation from both the current estimate $\overline{x}_{t}$ and the input image $I_{0}$.

Then Eq \ref{eq: Re(n)semble-DDPM_simple_minimize_term} can be modified to Eq \ref{eq: Re(n)semble-DDPM_simple_minimize_term_with_It}.

\begin{equation}
	\label{eq: Re(n)semble-DDPM_simple_minimize_term_with_It}
   \mathbb{E}_{x_{0},\epsilon,t}[||\epsilon - \epsilon_{\theta}(\overline{x}_{t}, I_{0}, t)||^{2}]
\end{equation}


During the training process, the end-to-end model is treated as a well-trained learner with its weights frozen. The denoising diffusion model is considered as the second learner that we need to train. The training pipeline of \ours is illustrated in Fig \ref{fig: pipeline_train} and detailed in Alg \ref{alg: Training Algorithm for Re(n)semble-DDPM}. We generate the final output $x_0$ by iteratively generating $\overline{x}_{0}$ through a reverse denoising process, and adding it with $\hat{x}_{0}$. The inference pipeline of \ours is illustrated in Fig \ref{fig: pipeline_inference} and detailed in Alg \ref{alg: Inference Algorithm for Re(n)semble-DDPM}. The specific working principle is described in Appendix \ref{sec: Working principle}. \par

\begin{algorithm}[t]
	\caption{Training Algorithm for \ours}
	\label{alg: Training Algorithm for Re(n)semble-DDPM}
	\begin{algorithmic}
		\REQUIRE total diffusion steps $T$, images and segmentation masks dataset $D={(I_{0}^n, x_{0}^n)}_{n}^{N}$, pretrained end-to-end model $E2EModel$.
		\STATE $\beta_{t}=\frac{10^{-4}(T-t)+2 \times 10^{-2}(t-1)}{T-1}$\\
		\STATE $\alpha_{t}=1-\beta_{t} $
		\STATE $\overline{\alpha}_{t}=  {\textstyle \prod_{s=1}^{t}} \alpha_{s}$
		\REPEAT
		\STATE Sample $(I_{0}^i, x_{0}^i)\sim D,\epsilon\sim N(0,I_{n \times n})$
		\STATE Sample $t\sim Uniform({1,...,T})$
		\STATE $\hat{x}_{0} = E2EModel(I_{0}) $  \COMMENT{no gradient and can be cached}
        \STATE $R = \hat{x}_{0} - x_{0}$ 
        \STATE $\overline{x}_{0} = x_{0} - R$ 
        \STATE $\overline{x}_{t}= \sqrt{\overline{\alpha}_{t}}\overline{x}_{0}+\sqrt{1-\overline{\alpha}_{t}}\epsilon $
        \STATE take gradient step on 
        \STATE \quad$\nabla_{\theta} ||\epsilon-\epsilon_{\theta}(\overline{x}_{t}, I_{0},t )||$
		\UNTIL{convergence}
	\end{algorithmic}
\end{algorithm}

\begin{algorithm}[t]
	\caption{Inference Algorithm for \ours}
	\label{alg: Inference Algorithm for Re(n)semble-DDPM}
	\begin{algorithmic}
		\REQUIRE total diffusion steps $T$, image $I_{0}$, pretrained end-to-end model $E2EModel$, pretrained \ours model $\epsilon_{\theta}$.
		\STATE $\beta_{t}=\frac{10^{-4}(T-t)+2 \times 10^{-2}(t-1)}{T-1}$
		\STATE $\alpha_{t}=1-\beta_{t} $
		\STATE $\overline{\alpha}_{t}=  {\textstyle \prod_{s=1}^{t}} \alpha_{s}$
        \STATE $\widetilde\beta_{t}=\frac{1-\overline{\alpha}_{t-1}}{1-\overline{\alpha}_{t}}\beta_{t}$
		\STATE Sample $\epsilon\sim N(0,I_{n \times n})$
		\STATE $\hat{x}_{0} = E2EModel(I_{0}) $ 
		\STATE $\overline{x}_{T} = \epsilon$
		\FOR{$t=T, T-1,...,2$}
		\STATE Sample $z\sim N(0,I_{n \times n})$
        \STATE $\overline{x}_{t-1}=\frac{1}{\sqrt{\alpha_{t}}}(\overline{x}_{t}-\frac{\beta_{t}}{\sqrt{1-\overline\alpha_{t}}}(\epsilon_{\theta}(\overline{x}_{t}, I_{0},t)) + \sqrt{\widetilde{\beta}}z $
		\ENDFOR
        \STATE $\overline{x}_{0}=\frac{1}{\sqrt{\alpha_{1}}}(\overline{x}_{1}-\frac{\beta_{1}}{\sqrt{1-\overline\alpha_{1}}}\epsilon_{\theta}(\overline{x}_{1}, I_{0},1)) $
        \RETURN $\frac{1}{2}(\overline{x}_{0}+\hat{x}_{0})$
	\end{algorithmic}
\end{algorithm}

In particular, when $R$ has a clear semantic meaning, such as in image restoration task, the $E2EModel$ is not necessary. We can rewrite $R$ as Eq \ref{eq: clear_R} and train \ours from scratch.

\begin{equation}
	\label{eq: clear_R}
	R = I_{0} - x_{0}
\end{equation}

Following \cite{ho2020denoising} we set

\begin{equation}
	\label{eq: Sigma_theta}
	\Sigma_{\theta}(x_{t},t)=\sigma_{t}^{2}I_{n \times n}
\end{equation}

where

\begin{equation}
	\label{eq: widetilde_beta}
	\widetilde\beta_{t}=\frac{1-\overline{\alpha}_{t-1}}{1-\overline{\alpha}_{t}}\beta_{t}
\end{equation}

\begin{equation}
	\label{eq: sigma_{t}^{2}}
	\sigma_{t}^{2}=\widetilde\beta_{t}
\end{equation}

The forward process variance parameter is a linearly increasing constant from $\beta_{1} = 10^{-4}$ to $\beta_{T} = 2 \times 10^{-2}$. $\beta_{t}$ can be formally represented as Eq \ref{eq: beta_t}.

\begin{equation}
	\label{eq: beta_t}
	\beta_{t}=\frac{10^{-4}(T-t)+2 \times 10^{-2}(t-1)}{T-1} 
\end{equation}

\section{Experiments}
\label{sec:experiments}

Detailed comparative experiments and ablation experiments on image segmentation will be released. Detailed comparative experiments in the fields of image denoising, image restoration, image dehazing, and image deraining will also be released. Detailed scale experiments will be released.

\section{Conclusion}
\label{sec:conclusion}

We propose \ours, a target image generation framework integrating denoising diffusion probabilistic models and existing end-to-end models for segmentation tasks by the strategy of ensemble learning. \par

By defining the residual term between the likelihood output distribution of the end-to-end model and the ground truth distribution, we propose a residual denoising diffusion probabilistic model by adding a new distribution learned through a Markov process. This new distribution is the sum of the ground truth distribution and the residual term, and it is strictly symmetric to the likelihood output distribution with respect to the ground truth.
During the training process, we freeze the weights of the end-to-end model, and only train the denoising diffusion probabilistic model.\par

We employ an ensemble learning strategy during the inference process. Since the output of the diffusion learner is designed to be symmetric with respect to the ground truth, the residual can be seamlessly eliminated, following the ensemble strategy derived as Eq \ref{eq: x_0}. \ours essentially narrows the gap between the likelihood output of the end-to-end model and the ground truth, thus further improving the performance.\par

Experimental results demonstrate that our model achieves good performance on various segmentation tasks and enhances the capability of the end-to-end model.Moreover, \ours is not only restricted to the field of image segmentation but can be generalized to any image generation tasks, such as image denoising, image super-resolution enhancement, and image restoration tasks. Furthermore, when the residual term has a clear semantic meaning and can be directly defined, the end-to-end model is not necessary, further supporting the generalization capability of \ours.\par

{
    \small
    \bibliographystyle{ieeenat_fullname}
    \bibliography{main}
}

\onecolumn
\appendix \section{Appendix Section}
\subsection{Working principle}
\label{sec: Working principle}

\begin{figure*}[h]
		\centering
		\includegraphics[width=0.85\linewidth]{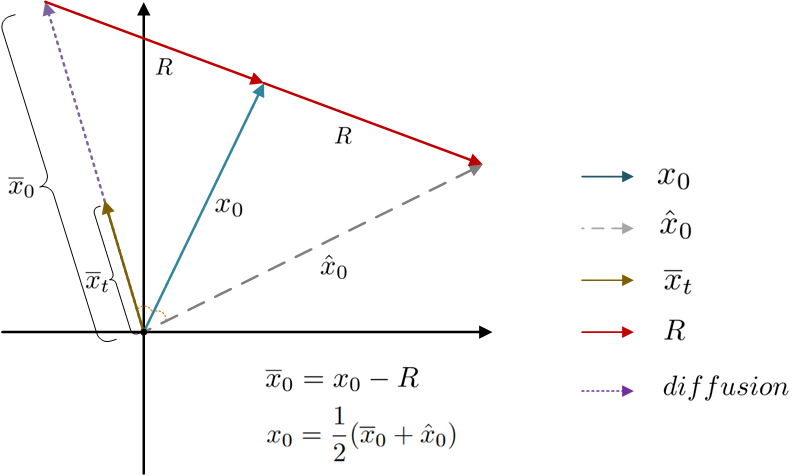}
		\caption{working principle of \ours.}
		\label{fig: working principle}
\end{figure*}
$R$ is defined by Eq \ref{eq: residual term} as $\hat{x}_{0}-x_0$, representing the residual term between the likelihood output of $E2EModel$ and the ground truth.  $\overline{x_0}$ is our learning target, defined by Eq \ref{eq: overline_x_0} as $x_0-R$. Both the $E2EModel$ sampling and our \ours sampling have implicitly learned the residual term $R$ along the opposite direction, and are symmetric with respect to the ground truth $x_0$. Based on this, the sampling spaces of two can be integrated to obtain a prediction for $x_0$, denoted as $\frac{1}{2}(\overline{x}_{0}+\hat{x}_{0})$ through Eq \ref{eq: x_0}. It mitigates the distribution distance between the sampling space of the $E2EModel$ and the ground truth by reducing the residual term. 

\end{document}